\documentclass[10pt,journal,compsoc]{IEEEtran}
\usepackage[nocompress]{cite} 

\usepackage[breaklinks=true,colorlinks,citecolor=blue,linkcolor=blue,urlcolor=blue]{hyperref}
\usepackage{url}            

\usepackage{subfigure}
\usepackage{color,colortbl}
\usepackage{multirow}
\usepackage{overpic}

\usepackage{times}
\usepackage{graphicx}
\usepackage{amssymb}

\usepackage{url}            
\usepackage{booktabs}       
\usepackage{nicefrac}       
\usepackage{microtype}      

\usepackage{cancel}  
\usepackage[normalem]{ulem}
\usepackage{mathtools}

\usepackage{url}

\usepackage{booktabs}       
\usepackage{xcolor}         

\usepackage{color}
\usepackage{arydshln}

\usepackage{etoolbox}
\makeatletter \patchcmd{\@makecaption}{\\}{.\  \justifying }{}{}\makeatother

\usepackage{ragged2e}

\usepackage{cancel}  
\usepackage[normalem]{ulem}

\usepackage{silence}
\newcommand{\myPara}[1]{\vspace{.1in} \noindent\textbf{#1}}

\def\ie{\emph{i.e.,~}}
\def\eg{\emph{e.g.,~}}

\def\Real{\mathbb{R}}
\newcommand{\figref}[1]{Fig.~\ref{#1}}%
\newcommand{\tabref}[1]{Table~\ref{#1}}%
\newcommand{\secref}[1]{Section~\ref{#1}}

\graphicspath{{./figures/}}

\def\Real{\mathbb{R}}

\begin{document}

\title{MDTv2: Masked Diffusion Transformer is a Strong Image Synthesizer}

\author{%
  Shanghua Gao, Pan Zhou, Ming-Ming Cheng, and Shuicheng Yan
  \IEEEcompsocitemizethanks{
  \IEEEcompsocthanksitem S. Gao and M.M. Cheng are with CS, Nankai University.
  \IEEEcompsocthanksitem Z. Pan is with Sea AI Lab.
  \IEEEcompsocthanksitem S. Yan is with Skywork AI.
  \IEEEcompsocthanksitem Part of this work was done while S. Gao was a research intern at Sea AI Lab.
  \IEEEcompsocthanksitem  Pan Zhou and M.M. Cheng are joint corresponding authors. 
		cmm@nankai.edu.cn.
  \IEEEcompsocthanksitem A preliminary version of this work has been presented in the ICCV 2023~\cite{Gao_2023_ICCV}.
  }
}

\IEEEtitleabstractindextext{\begin{abstract} \justifying
Despite its success in image synthesis, we observe that diffusion probabilistic models (DPMs) often lack contextual reasoning ability to learn the relations among object parts in an image, leading to a slow learning process. 
To solve this issue, we propose a Masked Diffusion Transformer (MDT) that introduces a mask latent modeling scheme to explicitly enhance the DPMs' ability to contextual relation learning among object semantic parts in an image. 
During training, MDT operates in the latent space to mask certain tokens.
Then, an asymmetric diffusion transformer is designed to predict masked tokens from unmasked ones while maintaining the diffusion generation process.
Our MDT can reconstruct the full information of an image from its incomplete contextual input, thus enabling it to learn the associated relations among image tokens. 
We further improve MDT with a more efficient macro network structure and
training strategy, named MDTv2. 
Experimental results show that MDTv2 achieves superior image synthesis performance, \eg a new SOTA FID score of 1.58 on the ImageNet dataset, and has more than 10$\times$ faster learning speed than the previous SOTA DiT.
The source code is released at \url{https://github.com/sail-sg/MDT}.
\end{abstract}
\begin{IEEEkeywords}
  Masked diffusion transformer, image generation
\end{IEEEkeywords}
}

\maketitle
\IEEEdisplaynontitleabstractindextext
\IEEEpeerreviewmaketitle

\IEEEraisesectionheading{\section{Introduction}\label{sec:introduction}}
\IEEEPARstart{D}{iffusion}
probabilistic models (DPMs)~\cite{dhariwal2021diffusion,rombach2022high} have been at the forefront of recent advances in image-level generative models, often surpassing the previously state-of-the-art (SOTA) generative adversarial networks (GANs) \cite{brock2018large,wu2019logan,razavi2019generating,gu2022rethinking}. 
Additionally, DPMs have demonstrated their success in numerous other applications, including text-to-image generation~\cite{rombach2022high}, image editing~\cite{gao2023editanything}, and speech generation~\cite{jeong2021diff}.   
DPMs adopt a time-inverted Stochastic Differential Equation (SDE) to gradually map a Gaussian noise into a sample by multiple time steps, with each step corresponding to a network evaluation.
In practice, generating a sample is time-consuming due to the thousands of time steps required for the SDE to converge.  
To address this issue, various generation sampling strategies \cite{ho2020denoising,lu2022dpm,salimans2022progressive} have been proposed to accelerate the inference speed. 
Nevertheless, improving the training speed of DPMs is less explored but highly desired. 
Training of DPMs also unavoidably requires a large number of time steps to ensure the convergence of SDEs, making it very computationally expensive, especially in this era where large-scale models \cite{dhariwal2021diffusion,peebles2022scalable} and data~\cite{imagenet_cvpr09,schuhmann2022laion,gao2021luss} are often used to improve generation performance. 

\newcommand{\addImg}[2]{\includegraphics[width=0.189\linewidth]{#1/sample304_#2}}
\newcommand{\addImgs}[1]{\addImg{#1}{05} &\addImg{#1}{1} &\addImg{#1}{2} &\addImg{#1}{3} &\addImg{#1}{30}}
\newcommand{\rotT}[1]{\rotatebox{90}{\small \qquad #1}}

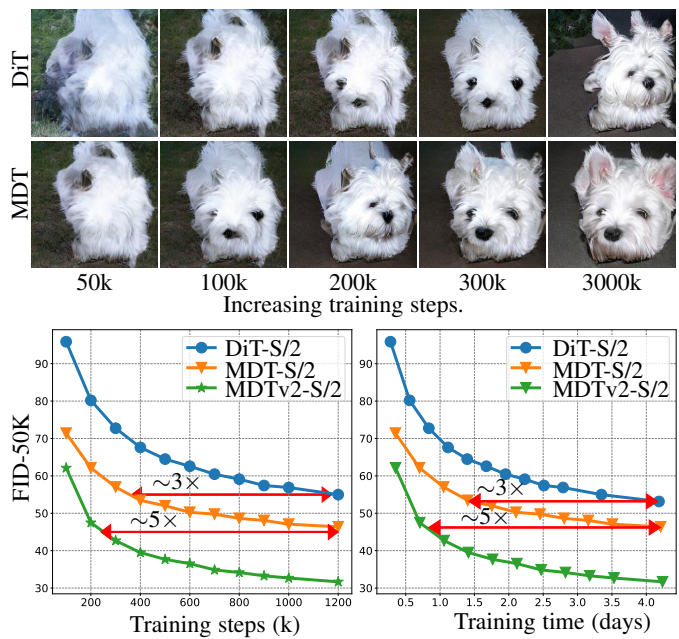
\begin{figure}[t]
  \small
  \raggedleft
  \setlength{\tabcolsep}{0.2mm}
  \renewcommand{\arraystretch}{0.7}
  \begin{tabular}{lccccc}
    \rotT{DiT} & \addImgs{dit} \\
    \rotT{MDT} & \addImgs{mdt} \\
    & 50k & 100k & 200k & 300k & 3000k \\
    \multicolumn{6}{c}{Increasing training steps.}
  \end{tabular} \\ \vspace{5pt}
  \begin{overpic}[width=0.475\linewidth]{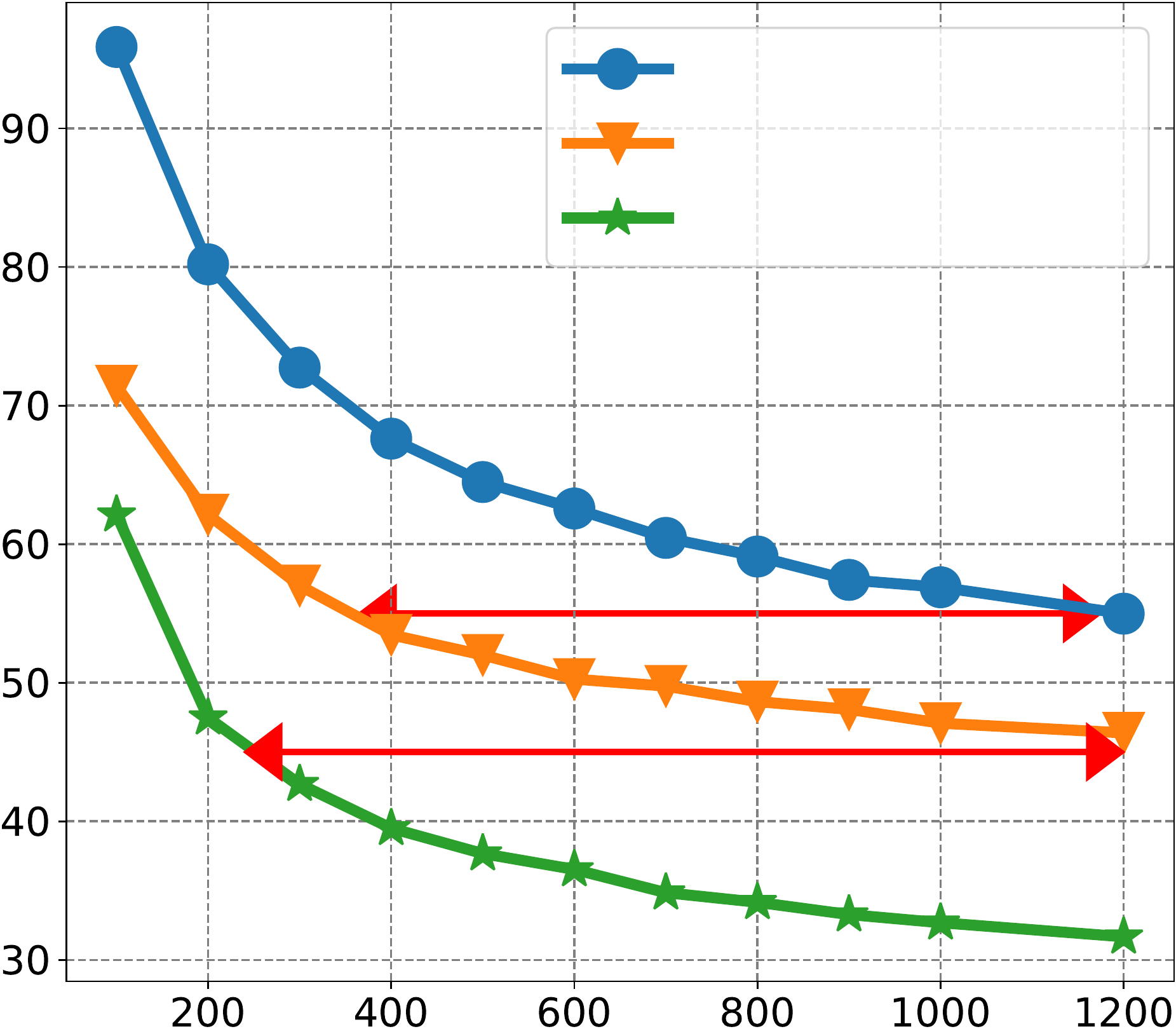}
    \put(60,79){DiT-S/2}
    \put(60,72){MDT-S/2}
    \put(60,66){MDTv2-S/2}
    \put(-7,18){\rotT{FID-50K}}
    \put(30,-8){Training steps (k)}
    \put(37,37){$\sim$3$\times$}
    \put(30,25){$\sim$5$\times$}
  \end{overpic}\vspace{3pt}
  \begin{overpic}[width=0.475\linewidth]{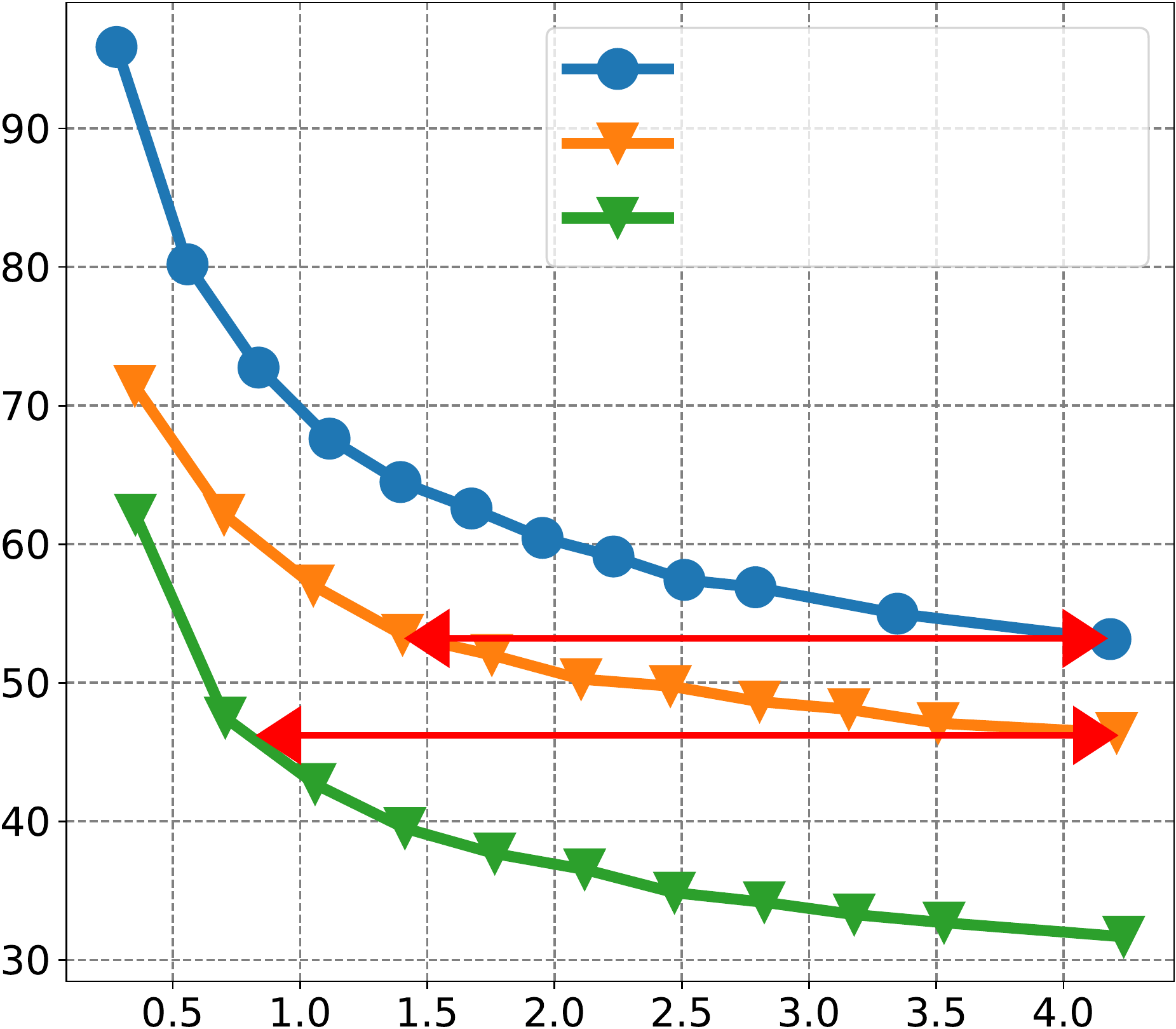} 
    \put(61,79){DiT-S/2}
    \put(61,72){MDT-S/2}
    \put(61,66){MDTv2-S/2}
    \put(30,-7){Training time (days)}
    \put(37,35){$\sim$3$\times$}
    \put(32,26){$\sim$5$\times$}
  \end{overpic}
  \caption{Top: Visual examples of MDT/DiT~\cite{peebles2022scalable}.
    Down: learning progress comparison between DiT, MDT, and MDTv2 w.r.t. training steps/time on 8$\times$A100 GPUs.
    MDT has about 3$\times$ faster learning speed than DiT while achieving superior FID scores. 
    MDTv2 further improves the training speed by about 5$\times$ compared to MDT.
  }\label{fig:converge_comp}
\end{figure}
  
In this work, we first observe that DPMs often struggle to learn the associated relations among object parts in an image. 
This leads to its slow learning process during training.   
Specifically, as illustrated in~\figref{fig:converge_comp}, the classical DPM, DDPM~\cite{ho2020denoising} with DiT~\cite{peebles2022scalable} as the backbone, has learned the shape of a dog at the 50k-th training step, then learns its one eye and mouth until at the 200k-th step while missing another eye.
Also, the relative position of two ears is not very accurate, even at the 300k-th step. %
This learning process reveals that DPMs fail to learn the associated relations among semantic parts and independently learn each semantic part.
This phenomenon is because DPMs maximize the log probability of real data by minimizing the per-pixel prediction loss, which ignores the associated relations among object parts in an image, thus resulting in their slow learning progress.

Inspired by the above observation, we propose an effective Masked Diffusion Transformer (MDT) to improve the training efficiency of DPMs.
MDT proposes a mask latent modeling scheme designed for transformer-based DPMs to explicitly enhance contextual learning ability and improve the associated relation learning among semantic parts in an image.  
Specifically, following~\cite{rombach2022high,peebles2022scalable}, MDT operates the diffusion process in the latent space to save computational costs. 
It masks certain image tokens and designs an asymmetric diffusion transformer structure to predict masked tokens from unmasked ones in a diffusion generation manner.
To this end, the asymmetric structure contains an encoder, a side-interpolater, and a decoder.  
The encoder and decoder modify the transformer block in DiT \cite{peebles2022scalable} by inserting global and local token position information to help predict masked tokens.
The encoder only processes remaining unmasked tokens during training, handling all tokens during inference, as there are no masks. 
To ensure the decoder always processes all tokens for training prediction or inference generation, a side-interpolater implemented by a small network aims to predict masked tokens from encoder output during training, and it is removed during inference.

With the masking latent modeling scheme, our MDT can reconstruct the full information of an image from its contextual incomplete input, learning the associated relations among semantic parts in an image. 
As shown in \figref{fig:converge_comp}, MDT typically generates two eyes of the dog at almost the same training steps, indicating that it correctly learns the associated semantics of an image by utilizing the mask latent modeling scheme.
In contrast, DiT~\cite{peebles2022scalable} cannot easily synthesize a dog with the correct semantic part relations. 
This comparison shows MDT's superior relation modeling and faster learning ability over DiT.  
Experimental results demonstrate that MDT achieves superior performance on the image synthesis task and sets the new SOTA on class-conditional image synthesis on the ImageNet dataset, as shown in~\figref{fig:sample} and \tabref{tab:SOTA_comp}.
MDT also enjoys about 3$\times$ faster learning progress during training than the SOTA DPMs, namely DiT, as demonstrated by \figref{fig:converge_comp} and \tabref{tab:ditvsmdt}. 
We hope our work can inspire more work on speeding up the diffusion 
training process with unified representation learning.

The main contributions are summarised as follows:
\begin{itemize}
\item By introducing an effective mask latent modeling scheme, we proposed a masked diffusion transformer method, which, for the first time, explicitly enhances the contextual learning ability of DPMs. 
\item Experiments show that our method better synthesizes images with much less training time than SOTA.
\end{itemize}

Compared to the ICCV 2023 version~\cite{Gao_2023_ICCV}, this journal version introduces a more efficient macro network structure for the masked diffusion transformer. 
This includes the integration of long shortcuts in the encoder~\cite{bao2022all} and dense input shortcuts in the decoder, aimed at facilitating learning in both the diffusion and masking reconstruction processes.
Furthermore, we have implemented enhanced training strategies. These include the adoption of a faster Adan optimizer~\cite{xie2022adan}, timesteps-adapted loss weights~\cite{Hang_2023_ICCV}, and expanding the masking ratio.
As a result of these improvements, MDTv2 demonstrates a learning speed that is approximately five times faster than the MDT.
We also introduce more details of the MDT, \eg the improved classifier-free guidance, network configuration details, and model complexity.
Additionally, we have conducted a more thorough analysis of MDT. 
This includes examining the position of the side-interpolator and the convergence speed, among other aspects. 
This comprehensive analysis provides deeper insights into the functionality and efficiency of MDT.

\newcommand{\addvis}[1]{\includegraphics[width=0.246\linewidth]{vis/#1.jpg}}
  
\begin{figure}[t]
  \centering
  \setlength{\tabcolsep}{0.1mm}
  \renewcommand{\arraystretch}{0.1}
  \begin{tabular}{ccccccccc}
    \addvis{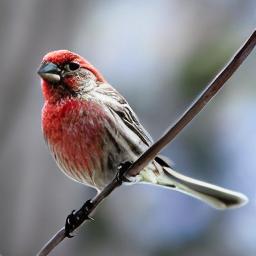} & \addvis{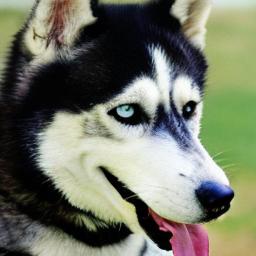} & 
    \addvis{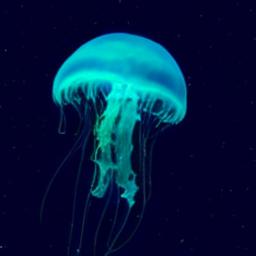} & \addvis{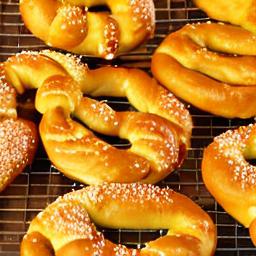}  \\
    \addvis{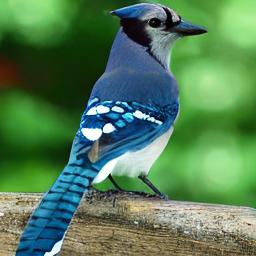}  & \addvis{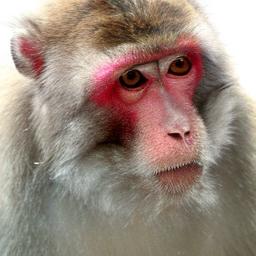}& 
    \addvis{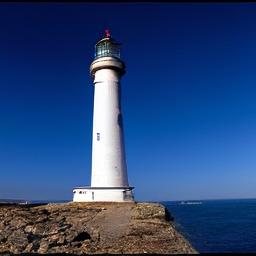} & \addvis{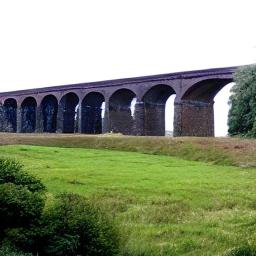}  \\
    \addvis{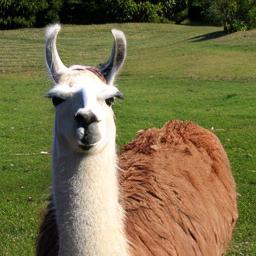} & \addvis{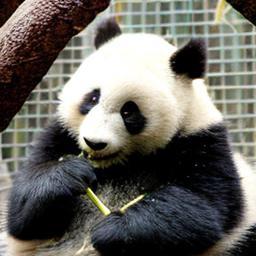} & 
    \addvis{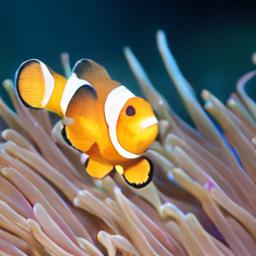} & \addvis{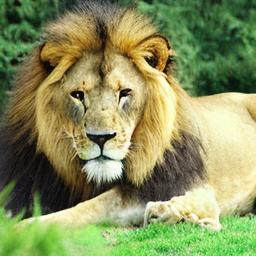}  \\
  \end{tabular} \\
  \caption{Visualization of images generated by the MDT-XL/2.}
  \label{fig:sample}
\end{figure}

\section{Related works}

\subsection{Diffusion Probabilistic models}

Diffusion probabilistic model (DPM)~\cite{ho2020denoising,dhariwal2021diffusion}, also known as score-based model~\cite{song2019generative,song2020improved}, is a competitive image synthesis approach.
DPMs begin by using an evolving Stochastic Differential Equation (SDE) to gradually add Gaussian noise into real data, transforming a complex data distribution into a Gaussian distribution.  
Then, it adopts a time-inverted SDE to map a Gaussian noise gradually into a sample by multiple steps.
At each sampling time step, a network is utilized to generate the sample along the gradient of the log probability, also known as the score function~\cite{song2020score}.
The iterative nature of diffusion models can result in high training and inference costs. 
Efficient sampling strategies \cite{ho2020denoising,lu2022dpm,salimans2022progressive,ho2022classifier,song2020denoising}, latent space diffusion~\cite{rombach2022high,vahdat2021score}, and multi-resolution cascaded generation~\cite{ho2022cascaded} have been proposed to reduce the inference cost.
Additionally, some training schemes~\cite{dockhorn2021score,bao2022estimating} are introduced to improve the diffusion model training, \eg, approximate maximum likelihood training \cite{song2021maximum,nichol2021improved,kingma2021variational}, and training loss weighting~\cite{kim2021soft,Karras2022edm}.
We identify the lack of contextual modeling ability in diffusion models as a key deficiency factor for slow convergence.
To address this, we propose the mask latent modeling scheme to enhance the contextual representation of diffusion models, which is orthogonal to existing diffusion training schemes.

\subsection{Networks for Diffusion Models}

The UNet-like~\cite{ronneberger2015u} network, enhanced by spatial self-attention~\cite{salimans2017pixelcnn++,van2016conditional} and group normalization~\cite{wu2018group} is firstly used for diffusion models~\cite{ho2020denoising}.
Several design improvements, \eg adding more attention heads \cite{GuoCvm22AttentionSurvey}, BigGAN~\cite{brock2018large} residual block, and adaptive group normalization, are proposed in \cite{dhariwal2021diffusion} to further enhance the generation ability of the UNet.
Due to the broad applicability of transformer networks, several works have recently attempted to utilize the vision transformer (ViT) structure for diffusion models~\cite{yang2022your,bao2022all,peebles2022scalable}.
GenViT~\cite{yang2022your} demonstrates that ViT can generate images but performs inferior to UNet.
U-ViT~\cite{bao2022all} improves ViT by adding long-skip connections and convolutional layers, achieving competitive performance with that of UNet.
DiT~\cite{peebles2022scalable} verifies the scaling ability of ViT on large model sizes and feature resolutions.
Our MDT is orthogonal to these diffusion networks, focusing on contextual representation learning.
Moreover, the position-aware designs in MDT reveal that the mask latent modeling scheme benefits from a stronger diffusion network.
We will explore further to release the potential of these networks with MDT.

\subsection{Mask Modeling}

Mask modeling has been proven to be effective in both recognition learning \cite{devlin2018bert,he2022masked,gao2022towards,cheng2021task} and generative modeling~\cite{radford2018improving,chang2022maskgit}.
In the natural language processing (NLP) field, mask modeling was first introduced to enable representation pretraining~\cite{devlin2018bert,radford2018improving} and language generation~\cite{brown2020language,MIR-Masked}.
Subsequently, it also proved feasible for vision recognition~\cite{bao2021beit} and generation~\cite{zhang2021m6,chang2022maskgit,gu2022vector} tasks.
In vision recognition, pretraining schemes that utilize mask modeling enable good representation quality~\cite{zhou2021ibot}, scalability~\cite{he2022masked} and faster convergence~\cite{gao2022towards}.
In generative modeling, following the bi-directional generative modeling in NLP, MaskGIT~\cite{chang2022maskgit} and MUSE~\cite{chang2023muse} use the masked generative transformer to predict randomly masked image tokens for image generation.
Similarly, VQ-Diffusion~\cite{gu2022vector} presents a mask-replace diffusion strategy to generate images.
In contrast, our MDT aims to enhance the contextual representation of the denoising diffusion transformer~\cite{peebles2022scalable} with mask latent modeling.
This preserves the detail refinement ability of denoising diffusion models by maintaining the diffusion process during inference.
To ensure that the mask latent modeling in MDT focuses on representation learning instead of reconstruction, we propose an asymmetrical structure in mask modeling training.
As an extra benefit, it enables lower training costs than masked generative models because it skips the masked patches in training instead of replacing masked input patches with a mask token.

\section{Masked Diffusion Transformer}

\myPara{Revisit Diffusion Probabilistic Model.}
For diffusion probabilistic models~\cite{dhariwal2021diffusion,sohl2015deep}, such as DDPM~\cite{ho2020denoising} and DDIM~\cite{song2020denoising}, training involves a forward noising process and a reverse denoising process. 
In the forward noising process, Gaussian noise $\epsilon \sim \mathcal{N}(0, \mathbf{I})$ is gradually added to the real sample $x_{0}$ via a discrete SDE formulation \cite{song2020denoising}.
If the time step $t$ is large, $x_t$ would be a Gaussian noise. 
Similarly, the reverse denoising process is a discrete SDE that gradually maps a Gaussian noise into a sample. 
At each time step, given $x_t$, it predicts the next reverse step $p_{\theta}(x_{t-1}|x_t)$ via a network.
Following~\cite{nichol2021improved,peebles2022scalable}, the network is trained by optimizing the variational lower-bound $L_{\text{vlb}}$ of the log-likelihood $p_{\theta}(x_0)$ ~\cite{sohl2015deep}.
During inference, one can sample a Gaussian noise and then gradually reverse to a sample $x_0$.
Same as~\cite{nichol2021improved,peebles2022scalable}, we train the diffusion model conditioned with class label $c$, \ie $p_{\theta}(x_{t-1}|x_t, c)$.
By default, we use class-conditioned image generation in our experiments.

\begin{figure}[!t]
  \centering
  \includegraphics[width=\linewidth]{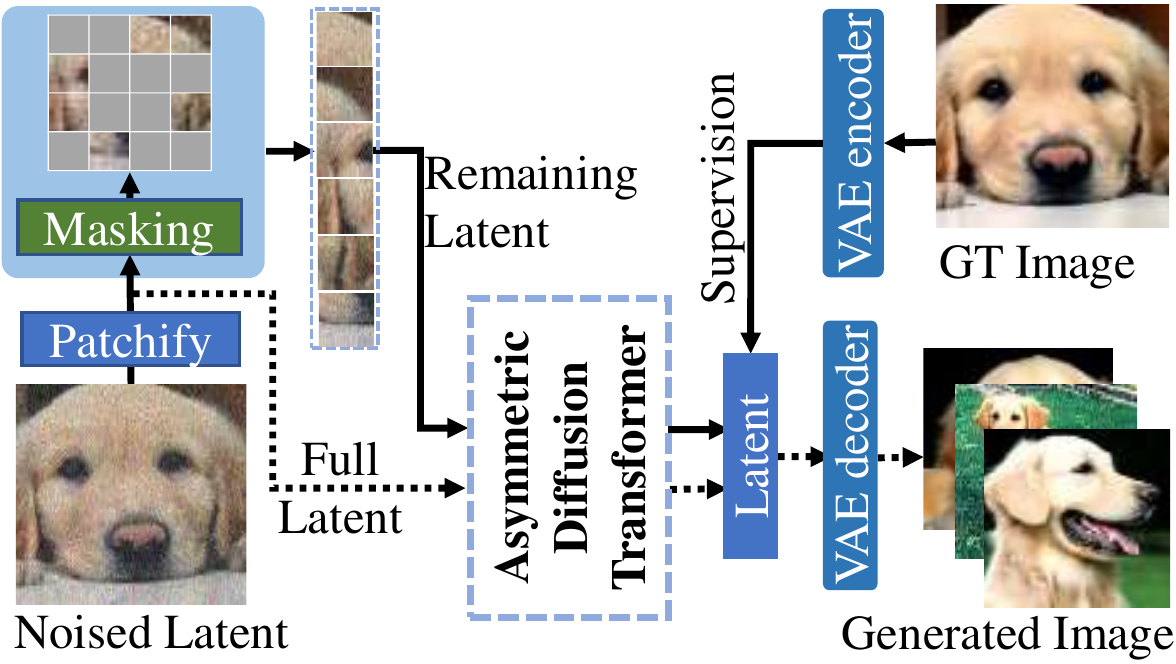}
  \caption{The overall framework of Masked Diffusion Transformer (MDT).
    Solid/dotted lines indicate each time step's training/inference process.
    Masking and side-interpolater are only used during training and are removed during inference. 
  }\label{fig:overall}
\end{figure}

\subsection{Overview}
\label{sec:mdt_method}

As shown in~\figref{fig:converge_comp}, DPMs with DiT backbone exhibit slow training convergence due to the slow learning of the associated relations among semantic parts in an image.
To relieve this issue, we propose a Masked Diffusion Transformer (MDT), which introduces a mask latent modeling scheme to enhance contextual learning ability explicitly.
To this end, as depicted in~\figref{fig:overall}, MDT consists of 1) a latent masking operation to mask the input image in the latent space, and 2) an asymmetric diffusion transformer structure that performs vanilla diffusion process as DPMs, but with masked input.
To reduce computational costs, MDT follows LatentDiffusion~\cite{rombach2022high} to perform generative learning in the latent space instead of raw pixel space.  

\begin{figure}[!t]
  \centering
  \small
  \begin{overpic}[width=\linewidth]{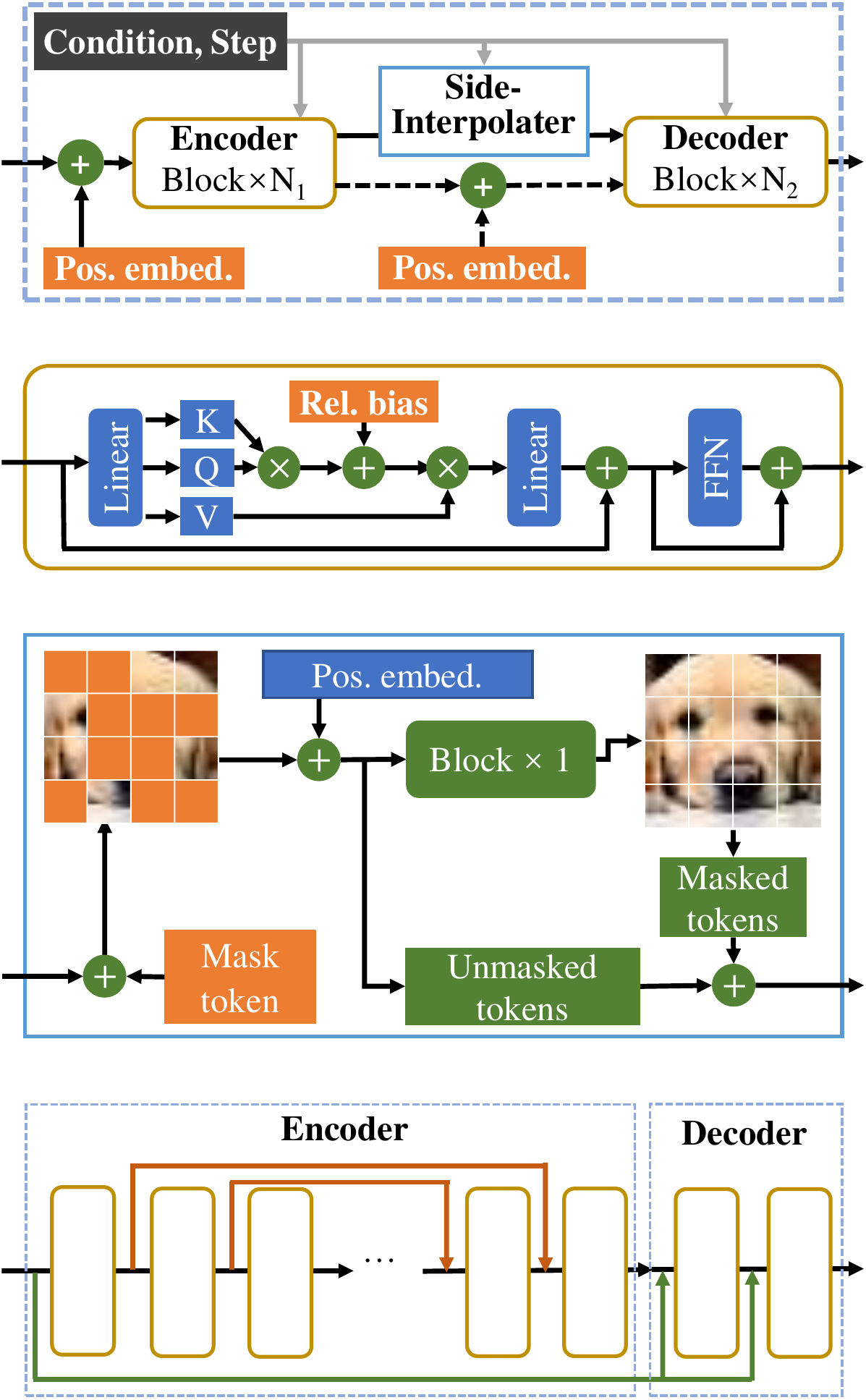}
    \put(9,76){(a) Asymmetric Diffusion Transformer}
    \put(18,57){(b) An Encoder/Decoder Block}
    \put(23,23.4){(c) Side-interpolater}
    \put(18,-2){(d) Macro-structure in MDTv2}
  \end{overpic} \\
  \caption{The asymmetric diffusion transformer structure in MDT.
    We modify the DiT~\cite{peebles2022scalable} by adding a side-interpolater, local relative positional bias, and learnable global position embeddings.
    Building upon MDT, MDTv2 introduces a macro-structure incorporating U-Net style long-shortcuts in the encoder and dense input-shortcuts in the decoder.
    The conditional scheme is omitted for simplicity.
  }\label{fig:block}
\end{figure}
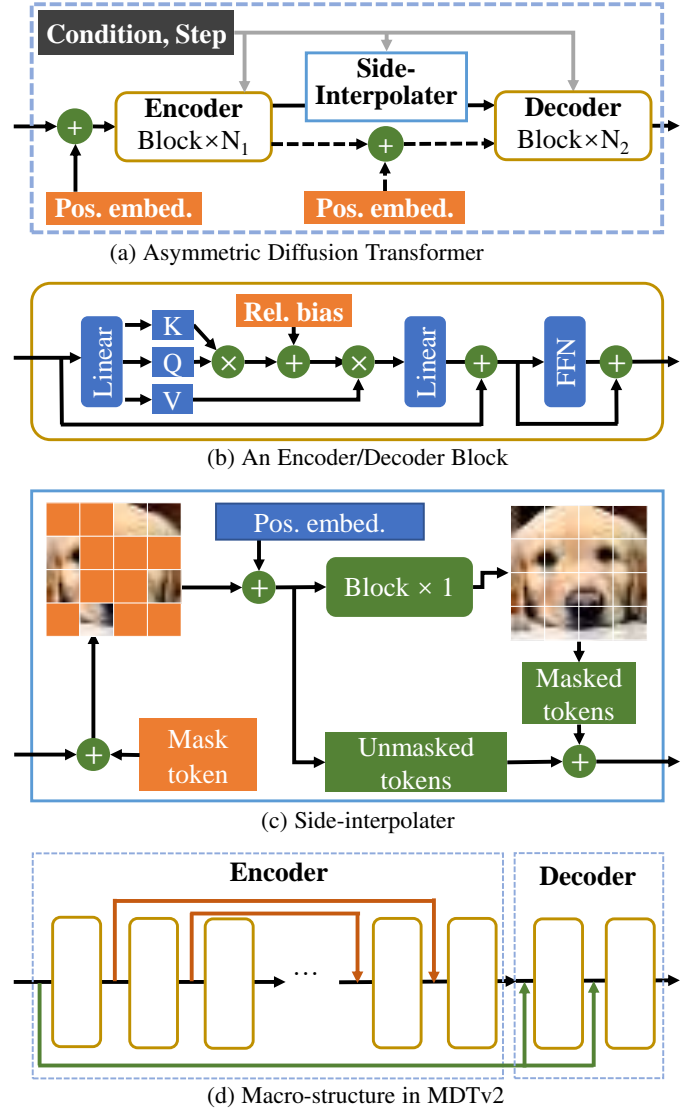

In the training phase, MDT first encodes an image into a latent space with a pre-trained VAE encoder~\cite{rombach2022high}. 
Then, MDT adds Gaussian noise into the image latent embedding.
The latent masking operation in MDT then patchifies the resulting noisy latent embedding into a sequence of tokens and masks certain tokens. 
The remaining unmasked tokens are fed into the asymmetric diffusion transformer, as shown in~\figref{fig:block}(a), which contains an encoder, a side-interpolater, and a decoder to predict the masked tokens from the unmasked ones.  
During inference, MDT replaces the side-interpolater with additional position embedding.
MDT takes the latent embedding of a Gaussian noise as input to generate the denoised latent embedding, which is then passed to a pre-trained VAE decoder~\cite{rombach2022high} for image generation.

The above masking latent modeling scheme in the training phase forces the diffusion model to reconstruct the full information of an image from its contextual incomplete input.
Thus, the model is encouraged to learn the relations among image latent tokens, particularly the associated relations among semantic parts in an image. 
For example, as illustrated in~\figref{fig:overall}, the model should first understand the correct associated relations among small image parts (tokens) of the dog image.
Then, it should generate the masked ``eye" tokens using remaining unmasked tokens as contextual information. 
Furthermore, \figref{fig:converge_comp} shows that MDT often learns to generate the associated semantics of an image at nearly the same pace, such as the generation of the dog's two eyes (two ears) at almost the same training step.
While DiT~\cite{peebles2022scalable} (DDPM with transformer backbone) learns to generate one eye (one ear) initially and then learns to generate another eye (ear) after roughly 100k training steps. 
This demonstrates the superior learning ability of MDT over DiT in terms of the associated relation learning of image semantics.

In the following parts, we will introduce the two key components of MDT, 
1) a latent masking operation,  
and 2) an asymmetric diffusion transformer structure.

\subsection{Latent Masking}	\label{sec:diff_train} 

Following the Latent Diffusion Model (LDM)~\cite{rombach2022high}, MDT performs generation learning in the latent space instead of raw pixel space to reduce computational costs. 
In the following, we briefly recall LDM and then introduce our latent masking operation on the latent input.

\myPara{Latent diffusion model (LDM).}
LDM employs a pre-trained VAE encoder $\textbf{E}$ to encode an image $v \in \Real^{3 \times H \times W}$ to a latent embedding $z = \textbf{E}(v) \in \Real^{c\times h \times w}$. 
It gradually adds noise to $z$ in the forward process 
and then denoises to predict $z$ in the reverse process.  
Finally, LDM uses a pre-trained VAE decoder $\textbf{D}$ to decode 
$z$ into a high-resolution image $v = \textbf{D}(z)$. 
Both the VAE encoder and decoder are fixed during training and inference.
Since $h$ and $w$ are smaller than $H$ and $W$, 
performing the diffusion process in the low-resolution latent space 
is more efficient than pixel space.

\myPara{Latent Masking Operation.} 
We first add Gaussian noise to an image's latent embedding $z$ during training. 
Following~\cite{peebles2022scalable}, we divide the noisy embedding $z$ into a sequence of $p\times p$-sized tokens and concatenate them to a matrix $u \in \Real^{d\times N}$, where $d$ is the channel number and $N$ is the number of tokens.  
Next, we randomly mask tokens with a ratio $\rho$ and concatenate the remaining tokens as $\hat{u} \in \Real^{d\times \hat{N}}$, where $\hat{N} = \rho N$.  
Accordingly, we can create a binary mask $M \in \Real^{N}$ in which one (zero) denotes the masked (unmasked) tokens.   
Finally, we feed the tokens $\hat{u}$ into our diffusion model for processing.  
We only use tokens $\hat{u}$ for two reasons. 
First, the model should focus on learning semantics instead of only predicting the masked tokens.
As shown in~\secref{sec:abl}, it achieves better performance than replacing the masked tokens with a learnable mask token and then processing all tokens like~\cite{bao2021beit,chang2022maskgit,chang2023muse}.
Second, it saves the training cost compared to processing all $N$ tokens.  
  
\subsection{Asymmetric Diffusion Transformer}
We introduce our asymmetric diffusion transformer for joint training of mask latent modeling and diffusion process. 
As shown in~\figref{fig:block}(a), it consists of three components: an encoder, a side-interpolater, and a decoder, each described in detail below. 
The term ``asymmetric" encompasses two distinct meanings: 1) the disparity between training and inference processes,
where the masked diffusion process is employed for training and the standard diffusion process is used during inference,
and 2) the network processes unmasked/full tokens in the encoder/decoder for the masked diffusion training.

\myPara{Position-aware encoder and decoder.} 
In MDT, predicting the masked latent tokens from the remaining unmasked tokens requires the position relations of all tokens.
To enhance the position information in the model, we propose a positional-aware encoder and decoder that facilitate the learning of the masked latent tokens.
Specifically, the encoder and decoder tailor the standard DiT block via adding two types of token position information containing $N_1$ and $N_2$ blocks, as illustrated in~\figref{fig:block}(a).

Firstly, the encoder adds the conventional learnable global position embedding into the noisy latent embedding input. 
Similarly, the decoder introduces the learnable position embedding into its input with different training and inference phase approaches. 
The side-interpolater already uses the learnable global position embedding introduced below during training.
During inference, since the side interpolater is discarded (see below), the decoder explicitly adds the position embedding into its input to enhance positional information.

Secondly, as depicted in~\figref{fig:block}(b), the encoder and decoder add a local relative positional bias~\cite{liu2021swin} to each head in each block when computing the attention score of the self-attention~\cite{vaswani2017attention}: 
\begin{equation*}
\mbox{Attention}(Q, K, V) = 
\mbox{Softmax}\left(\frac{QK^{\top}}{\sqrt{d_k}} + B_{r}\right) V,
\end{equation*}
where $Q$, $K$, and $V$, respectively denote the query, key, and value in the self-attention module, $d_k$ is the dimension of the key, and $B_{r} \in  \Real^{N \times N}$ is the relative positional bias.
$B_{r}$ is selected by the relative positional difference between the $i$-th position and other positions, updated during training. 
The local relative positional bias helps to capture the relative relations among tokens, facilitating the masking of latent modeling.

The encoder takes the unmasked noisy latent embedding provided by our latent masking operation and feeds its output into the side-interpolater/decoder during training/inference.   
The decoder's input is the output of the side-interpolater for training or the combination of the encoder output and the learnable position embedding for inference. 
During training, the encoder and decoder, respectively, handle remaining unmasked tokens and full tokens.
Thus, we name our model the ``asymmetric" model.  

\myPara{Side-interpolater.} 
As shown in \figref{fig:overall}, during training, for efficiency and better performance, the encoder only processes the remaining unmasked tokens $\hat{u}$.
While in the inference phase,  the encoder handles all tokens $u$ due to the lack of masks. 
This means that there is a big difference in the encoder output (\ie decoder input) during training and inference, at least in terms of token number. 
To ensure the decoder always processes all tokens for training prediction or inference generation, a side-interpolater implemented by a small network aims to predict masked tokens from encoder output during training and would be removed during inference.

In the training phase, the encoder processes the remaining unmasked tokens to obtain its output token embedding 
$\hat{q} \in \Real^{d\times \hat{N}}$.  
Then, as shown in~\figref{fig:block}(c), the side-interpolater first fills the masked positions, indicated by the mask $M$ defined in \secref{sec:diff_train}, by a shared learnable mask token, and also adds a learnable positional embedding to obtain an embedding 
$q \in \Real^{d\times N}$. 
Next, we use a basic encoder block to process $q$ to predict an interpolated embedding $\hat{k}$. 
The $\hat{k}$ tokens denote the predicted tokens. 
Finally,  we use a masked shortcut connection to combine prediction $\hat{k}$ and $q$ as $k = (1-M) \cdot q + M \cdot \hat{k}$.  
In summary, for masked tokens, we use the prediction by side-interpolater. 
For the remaining unmasked tokens, we still adopt the corresponding tokens in $q$. 
This can 
1) boost the consistency between training and inference phases, 
2) eliminates the mask-reconstruction process in the decoder.

Since there are no masks during inference, the side-interpolater is replaced by a position embedding operation, which adds the learnable position embeddings of the side-interpolater,
learned during training.  
This ensures the decoder always processes all tokens and uses the same learnable position embeddings for training prediction or inference generation, thus performing better image generation. 

\subsection{Masked Diffusion Transformer v2}
To further accelerate the diffusion training, MDTv2 incorporates a macro network structure based on the original masked diffusion transformer architecture, as illustrated in~\figref{fig:block}(d).
While the original MDT was modified based on DiT, featuring a plain network structure, MDTv2 introduces a macro network structure with enhanced shortcuts. 
This advancement significantly accelerates the convergence rate of MDT. 
Specifically, MDTv2 integrates U-Net-like long shortcuts in the encoder and dense input shortcuts in the decoder, further optimizing the overall architecture.

\myPara{Encoder with long-shortcuts.}
The U-Net style long-shortcut, \ie the long shortcut connections between shallow and deep layers of the network, has been proven effective in both ViT~\cite{bao2022all} and convolutional network~\cite{rombach2022high} based diffusion models.
We follow this design to enhance the encoder in MDT with U-Net style long-shortcuts.
Specifically, we consider an encoder composed of \(N_1\) blocks. 
The input and output of the block $i$ are denoted as $\mathsf{B}_i$ and $\hat{\mathsf{B}}_i$ respectively. 
As depicted in the red line of \figref{fig:block}(d), the input to block $i$ is defined as follows:
\begin{equation*}
\mathsf{B}_i =
\begin{cases}
\hat{\mathsf{B}}_{i-1}, & \text{if } 1 < i \leq \frac{N_1}{2}; \\
\hat{\mathsf{B}}_{i-1} \oplus \hat{\mathsf{B}}_{N_1-i+1}, & \text{if } N_1 \geq i > \frac{N_1}{2},
\end{cases}   
\end{equation*}
where \(\oplus\) represents the concatenation operation along the channel dimension.

\myPara{Decoder with dense input-shortcuts.}
An asymmetric structure is employed in the MDT framework to reduce training costs. 
Therefore, directly implementing long-shortcuts between the encoder and decoder is not feasible because masked encoder-layer patches are unavailable.  
Additionally, since the masked image patches are omitted in the encoder, the noise added to these masked patches is also lost. 
This loss of noise information hampers accurate noise prediction in the diffusion training process.
Considering the above, we propose the dense input-shortcuts that connect the input of the encoder 
to each block of the decoder.
Specifically,
as shown in the green line of \figref{fig:block}(d), 
given a decoder with \(N_2\) blocks,
the input of a decoder block $j$ where \(j \in [1, N_2]\), denoted as \(\mathsf{B}_j\), is defined as follows:
\begin{equation*}
\mathsf{B}_j  =
\hat{\mathsf{B}}_{j-1} \oplus u,
\end{equation*}
where $u$ is the noisy input as described in~\secref{sec:diff_train}.

In~\tabref{tab:structure_improvement}, we demonstrate that incorporating long-shortcuts in the encoder and dense input-shortcuts in the decoder significantly enhances the quality of image generation while reducing training time.

\subsection{Training and Inference}

\myPara{Training.}
During training, we feed both full latent embedding $u$ and the remaining latent embedding after masking $\hat{u}$ to the diffusion model.
We observe that only using the remaining unmasked latent embedding makes the model focus too much on masked region reconstruction,
while ignoring the diffusion training.
The training objectives for full/remaining unmasked latent inputs optimize the variational lower-bound, as in \cite{nichol2021improved,peebles2022scalable}.
Due to the asymmetrical masking structure, the extra costs for using the remaining latent embedding are small.  
This is also demonstrated by~\figref{fig:converge_comp}, which shows that MDT still achieves about 3$\times$ faster learning progress than previous SOTA DiT in total training hours. 

\myPara{Training strategy improvement in MDTv2.}
Compared with the MDT, we introduce new training strategies to enhance the convergence speed further. 
Specifically, the Adam optimizer is substituted with the Adan optimizer \cite{xie2022adan} to expedite the convergence process. 
Furthermore, to mitigate the conflicting optimization directions across timesteps, a Min-SNR weighting strategy is employed \cite{hang2023efficient}. 
This approach involves assigning loss weights for different timesteps based on the clamped signal-to-noise ratios.
Additionally, we find that employing a broader range of masking ratios, as opposed to a single fixed mask ratio in MDT, significantly enhances the model's capacity for contextual information modeling. 
Consequently, in MDTv2, the mask ratio is established within 30\%-5\%, diverging from the static 30\% mask ratio utilized in the MDT.
We demonstrate in~\tabref{tab:v1v2} that the aforementioned training strategies significantly enhance the convergence speed of MDT.

\begin{figure}[t]
  \raggedleft
  \vspace{3pt}
  \begin{overpic}[width=0.96\linewidth]{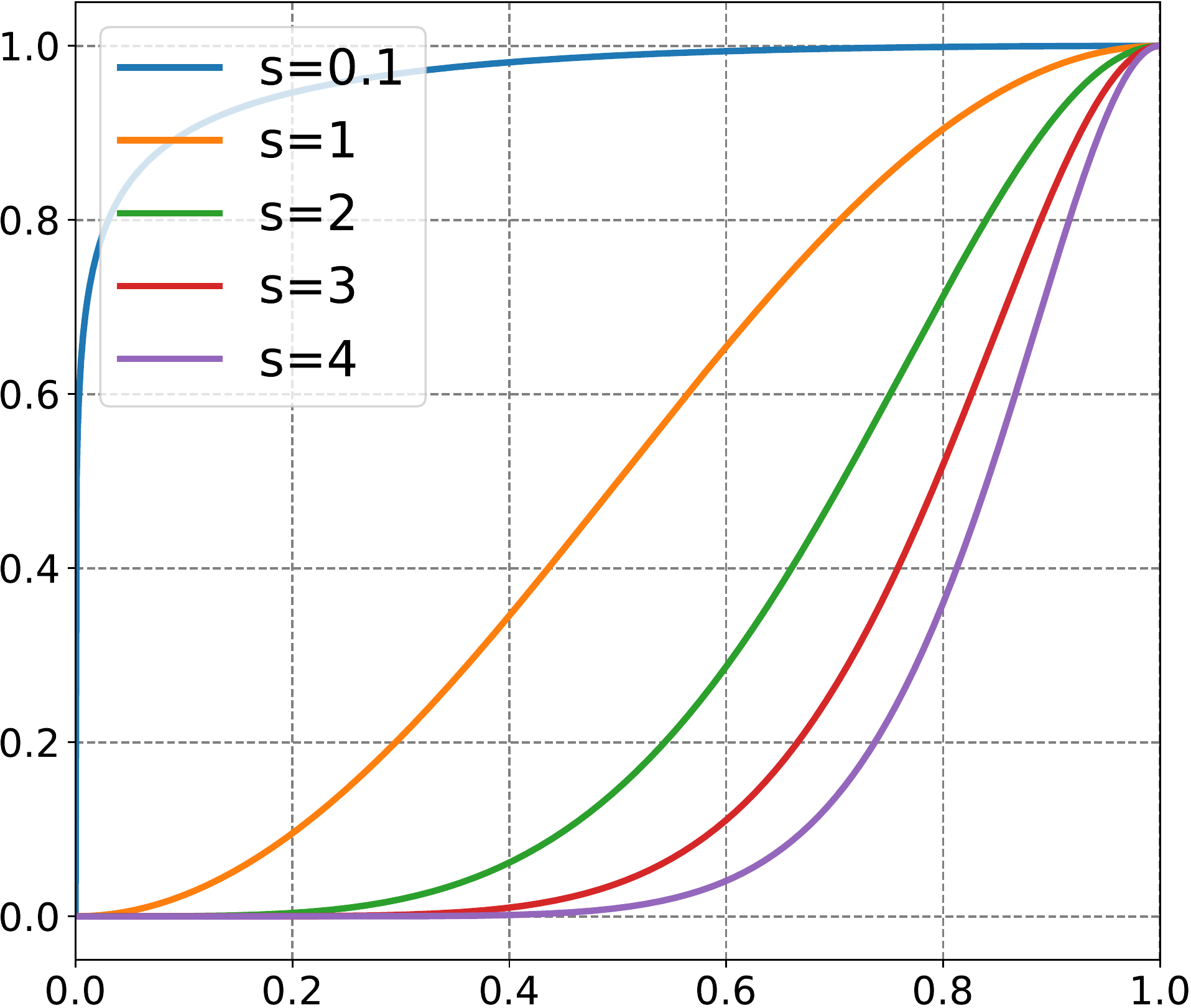}
    \put(-5,40){\rotatebox{90}{${w_t}/{w}$}}
    \put(48,-5){${t}/{t_{\text{max}}} $}
  \end{overpic}\vspace{8pt} \\
  \caption{The power-cosine scaling schedule for guidance scale in classifier-free guidance with difference $s$.
  A larger $s$ results in a slower increase of $w$ at early steps and a faster increase at late steps.
  }\label{fig:powercos}
\end{figure}

\myPara{Inference with improved classifier-free guidance.}
The classifier-free guidance sampling~\cite{ho2022classifier} enables
the trade-off between sample quality and diversity.
It achieves this by combining the class-conditional and unconditional
estimation:
\begin{equation*}
  \hat{\epsilon}_{\theta}(x_t, c) = 	\epsilon_{\theta}(x_t) + w \cdot ( \epsilon_{\theta}(x_t, c) - \epsilon_{\theta}(x_t) ),
\end{equation*}
where $\epsilon_{\theta}(x_t, c)$ is the class-conditional estimation,
$\epsilon_{\theta}(x_t)$ is the unconditional estimation, 
and $w$ is the guidance scale.
Generally, a larger $w$ results in high sample quality by decreasing the diversity.
MUSE~\cite{chang2023muse} changes the fixed guidance scale with a linear increasing schedule during sampling, which makes the model samples with more diversity at early steps while samples with higher fidelity at late steps.
Inspired by this, 
we present a power-cosine schedule for the guidance scale during the sampling procedure:
\begin{equation*}
  w_t = \frac{1-\cos{\pi \left(\frac{t}{t_{\text{max}}} \right)^s}}{2} w,
\end{equation*}
where $t$ is the time step during sampling, $t_{\text{max}}$ is the maximum sampling step, $w$ is the maximum guidance scale, and $s$ is a factor that controls the increasing speed of the guidance scale.
As revealed in~\figref{fig:powercos}, the power-cosine schedule enables a low guidance scale at early steps while quickly increasing the guidance scale at late steps.
By increasing $s$, the guidance scale has a slow increase at early steps and a fast increase at late steps.
The improved classifier-free guidance sampling equipped with the power-cosine guidance scale schedule enables the model samples with high diversity at early and high quality at late steps.
In this work, $s$ is set to 4, and the corresponding $w$ is set to 3.8 to ensure the model generates images with high fidelity at late steps.

\section{Experiments}\label{exp}

\subsection{Implementation}
We give the implementation details of MDT, including model architecture, training details, and evaluation metrics.

\subsubsection{Model Details}

\myPara{Model architecture.} 
We follow DiT~\cite{peebles2022scalable} to set the total block number (\ie, $N_1+N_2$), token number, and channel numbers of the diffusion transformer of MDT. 
As DiT reveals stronger synthesis performance when using a smaller patch size, we also use a patch size $p$=2 by default, denoted by MDT-/2. 
Moreover, We also follow DiT's parameters to design MDT for getting its small-, base-, and xlarge-sized model, denoted by MDT-S/B/XL.
The configurations of MDT models are given in~\tabref{tab:model_config}.
Same as LatentDiffusion~\cite{rombach2022high} and DiT, MDT adopts the fixed 
VAE\footnote{The model is downloaded in \url{https://huggingface.co/stabilityai/sd-vae-ft-mse}} provided by the Stable Diffusion to encode/decode the image/latent tokens by default.
The VAE encoder has a downsampling ratio of $1/8$, and a feature channel dimension of 4, \eg, an image of size 256$\times$256$\times$3 is encoded into a latent embedding of size 32$\times$32$\times$4.
The network parameters and training costs for MDT under different model scales are listed in~\tabref{tab:model_config}. 
In comparison to DiT baselines, MDT introduces a negligible extra inference
parameters and costs.
Additionally, MDTv2 has a similar model complexity compared to MDT.

\newcommand{\gr}[1]{\textcolor{gray}{#1}}
\begin{table}[t]
  \centering
  \small
  \setlength{\tabcolsep}{2mm} 
  \caption{Network configurations of MDT models.
  The configurations are following DiT networks~\cite{peebles2022scalable}.
  The layers consist of the layer numbers of encoder and decoder, and the decoder number $N_2$ is set to 2 for all models.
  FLOSs are measured with the latent embedding size of 32$\times$32 and $p$=2.
  The parameters and FLOSs are measured using the inference model.
  MDTv2 has similar model complexity compared to MDT.
  \label{tab:model_config}
  }
  \vspace{-3pt}
  \begin{tabular}{lcccccc}
    \toprule
    Size & Layers & Dim. & Head Num. & Param. (M) & FLOSs (G) \\	\midrule
    \multicolumn{6}{l}{Network configurations of MDT models.} \\ \hline
    S  & 12  & 384 & 6 & 33.1  & 6.07  \\               
    B  & 12  & 768 & 12 & 130.8  & 23.02 \\    
    XL & 28  & 1152 & 16 & 675.8 & 118.69  \\ \midrule \rowcolor{lightgray}
    \multicolumn{6}{l}{Network configurations of DiT baselines.} \\ \hline \rowcolor{lightgray}
    S & 12 & 384 & 6  & 32.9  & 6.06 \\  \rowcolor{lightgray}             
    B & 12 & 768 & 12 & 130.3 & 23.01 \\ \rowcolor{lightgray}   
    XL& 28 & 1152& 16 & 674.8 & 118.64 \\ 
    \bottomrule
  \end{tabular}
\end{table}

\myPara{Training details.}
Following~\cite{peebles2022scalable},
MDT models are trained by AdamW~\cite{loshchilov2017decoupled} optimizer
using 256 batch size, and without weight decay on ImageNet~\cite{imagenet_cvpr09} 
with an image resolution of 256$\times$256.
We set the mask ratio as 0.3 and $N_2=2$. 
Following the training settings in DiT,
we set the maximum step in training to 1000 and use the linear
variance schedule with a range from $10^{-4}$ to $2\times10^{-2}$.
Other settings are also aligned with DiT.
For MDTv2 models, the optimizer is changed to Adan~\cite{xie2022adan},
The mask ratio varies between 0.3 and 0.5,
and $N_2$ is increased to 4 for XL/L sizes models and increased to 6 for S size models.

\myPara{Evaluation.}
We evaluate models with commonly used metrics,
\ie Fre'chet Inception Distance (FID)~\cite{heusel2017gans},
sFID~\cite{nash2021generating}, Inception Score (IS)~\cite{salimans2016improved}, 
Precision and Recall~\cite{kynkaanniemi2019improved}.
The FID is used as the major metric as it measures both diversity and fidelity.
sFID improves upon FID by evaluating at the spatial level.
As a complement, IS and Precision are used for measuring fidelity,
and Recall is used to measure diversity.
For fair comparisons, we follow~\cite{peebles2022scalable}
to use the TensorFlow evaluation suite from ADM~\cite{dhariwal2021diffusion}
and report FID-50K with 250 DDPM sampling steps.
Unless specified otherwise,
we report the FID scores without the classifier-free guidance~\cite{ho2022classifier}.

\begin{table}[t]
  \centering
  \setlength{\tabcolsep}{2.4mm} 
  \small
  \caption{Comparison between DiT~\cite{peebles2022scalable} and MDT under 
    different model sizes and training steps on ImageNet 256$\times$256. 
    DiT results are obtained from DiT reported results.
  }\label{tab:ditvsmdt}
  \vspace{-8pt}
  \begin{tabular}{lcccccc}\toprule
    Method & Image Res. & Training Steps (k) & FID-50K$\downarrow$  \\	\midrule
    DiT-S/2 & 256$\times$256 & 400 & 68.40   \\ 
    \hdashline
    MDT-S/2 & 256$\times$256 & 300 & 57.01   \\
    MDT-S/2 & 256$\times$256 & 400 & 53.46   \\
    MDT-S/2 & 256$\times$256 & 2000 & 44.14  \\ 
    MDT-S/2 & 256$\times$256 & 3500 & 41.37  \\ \hdashline
    MDTv2-S/2 & 256$\times$256 & 400 & 39.50 \\
    MDTv2-S/2 & 256$\times$256 & 1500 & \textbf{31.05} \\
    \midrule
    DiT-B/2 & 256$\times$256 & 400 & 43.47   \\ \hdashline
    MDT-B/2 & 256$\times$256 & 400 & 34.33   \\
    MDT-B/2 & 256$\times$256 & 3500 & 20.45  \\ \hdashline
    MDTv2-B/2 & 256$\times$256 & 400 & 19.55  \\
    MDTv2-B/2 & 256$\times$256 & 1600 & \textbf{13.60}  \\
    \midrule
    DiT-XL/2 & 256$\times$256 & 400 & 19.47   \\
    DiT-XL/2  & 256$\times$256 & 2352 & 10.67   \\
    DiT-XL/2  & 256$\times$256 & 7000 & 9.62   \\
    \hdashline
    MDT-XL/2 & 256$\times$256 & 400 & 16.42    \\
    MDT-XL/2 & 256$\times$256 & 1300 & 9.60    \\
    MDT-XL/2 & 256$\times$256 & 6500 & 6.23 \\ 
  \hdashline
    MDTv2-XL/2 & 256$\times$256 & 400 & 7.70\\
    MDTv2-XL/2 & 256$\times$256 & 700 &  6.11\\
    MDTv2-XL/2 & 256$\times$256 & 1000 & 5.80\\
    MDTv2-XL/2 & 256$\times$256 & 2000 &  \textbf{5.06}\\
    \bottomrule
  \end{tabular}
\end{table}

\begin{table}
  \centering
  \setlength{\tabcolsep}{0.4mm} 
  \small
  \caption{Comparison with existing methods on class-conditional image generation with the ImageNet 256$\times$256 dataset. 
    -G denotes the results with classifier-free guidance~\cite{ho2022classifier}.
    Results of MDT-XL/2 model are given for comparison.
    Compared results are obtained from their papers.
  }
  \label{tab:SOTA_comp}
  \vspace{-8pt}
  \begin{tabular}{lcccccc} \toprule
    Method	& Cost(Iter$\times$BS) & FID$\downarrow$ & sFID$\downarrow$ & IS$\uparrow$ & Prec.$\uparrow$ & Rec.$\uparrow$ \\	\midrule
    DCTrans.\cite{nash2021generating} & - & 36.51 & - & - & 0.36 & 0.67 \\
    VQVAE-2\cite{razavi2019generating} & - & 31.11 & - & - & 0.36 & 0.57 \\
    VQGAN\cite{esser2021taming} & - & 15.78 & 78.3 & - & - & - \\
    BigGAN-deep\cite{brock2018large} & - & 6.95 & 7.36 & 171.4 & 0.87 & 0.28 \\
    StyleGAN\cite{sauer2022stylegan} & - & 2.30 & 4.02 & 265.12 & 0.78 & 0.53 \\
    Impr. DDPM\cite{nichol2021improved} & - & 12.26 & - & - & 0.70 & 0.62 \\
    MaskGIT\cite{chang2022maskgit} & 1387k$\times$256 & 6.18 & - & 182.1 & 0.80 & 0.51 \\
    CDM\cite{ho2022cascaded}  & - & 4.88  & - & 158.71 & - & - \\
    \midrule
    ADM\cite{dhariwal2021diffusion}  &1980k$\times$256& 10.94 & 6.02 & 100.98 & 0.69 & 0.63 \\
    LDM-8\cite{rombach2022high} & 4800k$\times$64 & 15.51 & - & 79.03 & 0.65 & 0.63 \\
    LDM-4 & 178k$\times$1200 & 10.56  & - & 103.49 & 0.71  & 0.62 \\
    \hdashline
    DiT-XL/2\cite{peebles2022scalable} & 7000k$\times$256  & 9.62  & 6.85  & 121.50  & 0.67  & 0.67 \\
    \textbf{MDT} & 2500k$\times$256  & 7.41  & 4.95 & 121.22 & 0.72 & 0.64 \\
    \textbf{MDT} & 6500k$\times$256  & 6.23  & 5.23 & 143.02 & 0.71 & 0.65 \\
    \textbf{MDTv2} & \textbf{2000k$\times$256} &  \textbf{5.06} & \textbf{4.56} & \textbf{155.58} &  \textbf{0.72} & 0.66 \\
    \midrule
    ADM-G\cite{dhariwal2021diffusion} &1980k$\times$256 & 4.59 & 5.25 & 186.70 & 0.82 & 0.52 \\
    ADM-G, U & 1980k$\times$256 & 3.94 & 6.14 & 215.84 & 0.83 & 0.53 \\
    LDM-8-G\cite{rombach2022high} & 4800k$\times$64 & 7.76 & - & 209.52 & 0.84 & 0.35 \\
    LDM-4-G & 178k$\times$1200 & 3.60 & - & 247.67  & 0.87  & 0.48 \\  
    U-ViT-G~\cite{bao2022all}     & 300k$\times$1024 & 3.40 & - & - & - & - \\
    \hdashline
    DiT-XL/2-G\cite{peebles2022scalable} & 7000k$\times$256 & 2.27 & 4.60 & 278.24 & 0.83 & 0.57 \\
    \textbf{MDT-G} & 2500k$\times$256  & 2.15  & 4.52 & 249.27 & 0.82 & 0.58 \\
    \textbf{MDT-G} & 6500k$\times$256  & 1.79  & 4.57 & 283.01 & 0.81 & 0.61 \\ 
    \textbf{MDTv2-G} & 3500k$\times$256  & 1.63  & \textbf{4.45} & 311.73 & 0.79 & 0.64 \\
    \textbf{MDTv2-G} & 4600k$\times$256  & \textbf{1.58}  & 4.52 & \textbf{314.73} & 0.79 & \textbf{0.65} \\
    \bottomrule
  \end{tabular}
\end{table}

\subsection{Comparison Results}

\myPara{Performance comparison.}
\tabref{tab:ditvsmdt}  compares our MDT with the SOTA DiT  
under different model sizes. 
It is evident that 
MDT achieves higher FID scores for all model scales with fewer training costs.
The parameters and inference cost of MDTs are similar to DiT, since 
  the extra modules in MDT are negligible as introduced in~\secref{sec:mdt_method}. 
For small models, MDT-S/2 trained with 300k steps outperforms the DiT-S/2
trained with 400k steps by a large margin on FID (57.01 vs. 68.40). More importantly,
MDT-S/2 trained with 2000k steps achieves similar performance with a larger model DiT-B/2 trained with a similar computational budget.
MDTv2-S/2 significantly surpasses both MDT and DiT-S/2, 
achieving better performance with fewer training steps. Specifically, when trained for 400k steps, MDTv2-S/2 outperforms MDT-S/2, which is trained for 3,500k steps, by an FID margin of 1.87, utilizing 9 times fewer training steps.
For the largest model,
MDT-XL/2 trained with 1300k steps outperforms
DiT-XL/2 trained with 7000k steps on FID (9.60 vs. 9.62),
achieving about 5$\times$ faster training progress.
Furthermore, MDTv2-XL/2, after being trained for 400k steps, significantly outperforms DiT-XL/2, which was trained for 7000k steps, by achieving a substantial improvement of 1.92 in FID. 
This indicates an impressive training acceleration of more than 18 times. 
Remarkably, MDTv2-XL/2 also achieves a 9 times faster training speed compared to the original MDT model, with FID scores of 6.11 versus 6.23 obtained after 700 versus 6500 training steps, respectively.

We also compare the class-conditional image generation performance of MDT with existing methods
in~\tabref{tab:SOTA_comp}. 
To make fair comparisons with DiT,
we also use the EMA weights of VAE decoder in this table.
Under class-conditional settings, MDT achieves superior performance to DiT with fewer training iterations,
achieving a FID of 6.23 compared to DiT's 9.62. 
Additionally, MDTv2, with under a third of MDT's training iterations, further improves efficiency, lowering the FID from 6.23 to 5.06.
Based previous works~\cite{rombach2022high,dhariwal2021diffusion,peebles2022scalable,bao2022all},
we propose an improved classifier-free guidance~\cite{ho2022classifier}
with a power-cosine weight scaling
to trade off between precision and recall
during class-conditional sampling.
MDT achieves superior performance over  
previous SOTA DiT and other methods with the FID score of 1.79,
setting a new SOTA for class-conditional image generation.
MDTv2 further advances this achievement, \emph{pushing the SOTA for class-conditional image generation to the FID score of 1.58 with fewer training iterations}.
Similar to DiT,
we never observe the model has saturated FID scores when continuing training.

\myPara{Convergence speed.}
\figref{fig:converge_comp} compares the performance of the DiT-S/2 baseline, MDT-S/2 and MDTv2-S/2 
under different training steps and training time on  8$\times$A100 GPUs.
Because of the stronger contextual learning ability,
MDT achieves better performance with faster generation learning speed.
MDT enjoys about 3$\times$ faster learning speed in terms of both training steps and training time. 
MDTv2 further enhances the training speed by about 5$\times$ compared to MDT.
For instance, when trained for approximately 33 hours (400k steps), MDT-S/2 demonstrates enhanced performance compared to DiT-S/2, which requires about 100 hours (1500k steps) of training. Furthermore, MDT${\text{v2}}$ reduces training time even more, to just 13 hours (15k steps).
This reveals that contextual learning is vital for
faster generation learning of diffusion models.

\begin{table*}
  \centering
  \caption{Ablation study on MDT-S/2. Models are trained for 600k iterations.}
  \vspace{-10pt}
  \subfigure[Effect of asymmetric masking structure.]{
    \centering
    \setlength{\tabcolsep}{7.1mm}
    \begin{tabular}{cc} \toprule
      Asymmetric stru. & FID-50k$\downarrow$ \\	\midrule
      $\times$     & 51.56  \\ 
      $\checkmark$ & \textbf{50.26}  \\ \bottomrule
    \end{tabular}
    \label{tab:asymmetric_mask}
  } \hfill
  \subfigure[Effect of side-interpolater.]{
    \centering
    \setlength{\tabcolsep}{5.4mm}
    \begin{tabular}{cc} \toprule
      Side-interpolater & FID-50k$\downarrow$ \\	\midrule
      $\times$     & 51.60  \\ 
      $\checkmark$ & \textbf{50.26}  \\  \bottomrule
    \end{tabular}
    \label{tab:side_interpolater}
  } \hfill
  \subfigure[Effect of masked shortcut.]{
    \centering
    \setlength{\tabcolsep}{5.4mm}
    \begin{tabular}{cc} \toprule
      Masked shortcut & FID-50k$\downarrow$ \\	\midrule
      $\times$     & 50.91  \\ 
      $\checkmark$ & \textbf{50.26}  \\  \bottomrule
    \end{tabular}
    \label{tab:masking_shortcut}
  } \\
  \subfigure[Full/remaining unmasked latent under aligned cost.]{
    \centering
    \setlength{\tabcolsep}{7.3mm}
    \begin{tabular}{cc} \toprule
      Latent type & FID-50k$\downarrow$ \\	\midrule
      Full+Unmasked & \textbf{50.26}  \\ 
      Full        & 52.30 \\ 
      Unmasked      & 76.63 \\ \bottomrule
    \end{tabular}
    \label{tab:used_latent}
  } \hfill
  \subfigure[Supervision on token parts.]{
    \centering
    \setlength{\tabcolsep}{7.5mm}
    \renewcommand{\arraystretch}{1.33}
    \begin{tabular}{cc} \toprule
      Sup. parts & FID-50k$\downarrow$ \\	\midrule
      All    & \textbf{50.26}  \\ 
      Masked & 58.35    \\ \bottomrule
    \end{tabular}
    \label{tab:mask_sup}
  } \hfill
  \subfigure[Number of blocks in side-interpolator.]{
    \centering
    \setlength{\tabcolsep}{8.1mm}
    \begin{tabular}{cc} \toprule
      Number & FID-50k$\downarrow$ \\	\midrule
      1 & \textbf{50.26}  \\ 
      2 & 51.77 \\ 
      3 & 51.96  \\ \bottomrule
    \end{tabular}
  \label{tab:num_block_si}
  } \\
  \subfigure[Effect of positional embeddings in SI.]{
    \centering
    \setlength{\tabcolsep}{7.8mm}
    \begin{tabular}{cc} \toprule
      IS Pos. embed. & FID-50k$\downarrow$ \\	\midrule
      $\times$     & 51.58  \\ 
      $\checkmark$ & \textbf{50.26}  \\  \bottomrule
    \end{tabular}
    \label{tab:pos_embed_si}
  } \hfill
  \subfigure[Effect of learnable positional embeddings.]{
    \centering
    \setlength{\tabcolsep}{6mm}
    \begin{tabular}{cc} \toprule
      Learnable pos. & FID-50k$\downarrow$ \\	\midrule
      $\times$     &     50.80  \\ 
      $\checkmark$ &     \textbf{50.26}  \\ \bottomrule
    \end{tabular}
    \label{tab:learn_pos}
  } \hfill
  \subfigure[Effect of relative positional bias.]{
    \centering
    \setlength{\tabcolsep}{5.2mm}
    \begin{tabular}{cc} \toprule
      Relative pos. bias & FID-50k$\downarrow$ \\	\midrule
      $\times$     & 53.56  \\ 
      $\checkmark$ & \textbf{50.26}  \\  \bottomrule
    \end{tabular}
  \label{tab:rel_bias}
  } \\
\end{table*}

\subsection{Ablation on MDT}
\label{sec:abl}
In this part, we conduct ablation to verify the designs in MDT.
We report the results of MDT-S/2 model and use FID-50k as the evaluation
metric unless otherwise stated.

\myPara{Asymmetric vs. Symmetric architecture in masking.}
Unlike the masked generation works~\cite{chang2022maskgit,chang2023muse},
\eg, MaskGIT,
that utilize the masking scheme to generate images,
MDT focuses on improving diffusion models with contextual learning ability  via
the masking latent modeling.
Therefore, we use an asymmetric architecture to only process the remaining unmasked tokens
in the diffusion model encoder.
We compare the asymmetric architecture in MDT and the symmetrical architecture~\cite{chang2022maskgit}
that processes full input with masked tokens replaced by a learnable mask token.
As shown in~\tabref{tab:asymmetric_mask},
the asymmetric architecture in MDT has an FID of 50.26, outperforming
  the FID of 51.56 
  achieved by the symmetric architecture.
The asymmetric architecture further reduces the training cost and allows  
the diffusion model to focus on learning contextual information instead of
reconstructing masked tokens.

\myPara{Effect of side-interpolater.}
The side-interpolater in MDT predicts the masked tokens,
allowing the diffusion model to learn more semantics
and maintain consistency in decoder inputs during training and inference.
We compare the performance with/without the side-interpolater in
\tabref{tab:side_interpolater},
and observe a gain of
1.34 in FID when using the side-interpolater,
proving its effectiveness.

\myPara{Masked shortcut in side-interpolater.}
The masked shortcut ensures that  the side-interpolater only predicts the
masked tokens from unmasked ones. 
\tabref{tab:masking_shortcut} shows
that using the masked shortcut enhances the FID
from 50.91 to 50.26,
indicating that restricting side-interpolater
to only predict masked tokens helps the diffusion model achieve stronger performance.

\myPara{Full and remaining unmasked latent tokens.}
In MDT, both the full and remaining unmasked latent embeddings
are fed into the diffusion model during training.
In comparison, we give the results trained by only using full/remaining unmasked latent embeddings
as shown in~\tabref{tab:used_latent},
where the computational cost is aligned for fair comparisons.
Trained with both full and remaining unmasked latent leads to clear gain
over two competitors.
While using only the remaining unmasked latent embeddings results in slow convergence,
which we attribute to the training/inference inconsistency
as the inference in MDT is a diffusion process instead of the masked reconstruction process.

\myPara{Loss on all tokens.}
By default, we calculate the loss on both masked and remaining unmasked latent embeddings.
In comparison, mask modeling for recognition models commonly 
calculates loss on masked tokens~\cite{he2022masked,bao2021beit}.
\tabref{tab:mask_sup} shows that
calculating the loss on all tokens is much better than on masked tokens.
We assume that this is because generative models require stronger consistency among patches than recognition models do,
since details are vital for high-quality image synthesis.

\begin{table}[t!]
  \centering
  \small
  \setlength{\tabcolsep}{2.3mm} 
  \caption{Effect of different masking ratios. 
    MDT-S/2 are trained with 600k iterations. 
  }\label{tab:maskratio}
  \vspace{-8pt}
\begin{tabular}{ccccccc}
    \toprule
    Mask Ratio	 &  FID$\downarrow$ & sFID$\downarrow$ & IS$\uparrow$ & Precision$\uparrow$ & Recall$\uparrow$ \\	\midrule
    0.1 & 51.60 & 10.23 & 26.65 & 0.44 & 0.60   \\
    0.2 & 51.44 & 10.09 & 26.75 & 0.44 & 0.58    \\
    \textbf{0.3} & \textbf{50.26} & \textbf{10.08} & \textbf{27.61} & \textbf{0.45} & 0.60   \\
    0.4 & 50.88 & 10.21 & 27.44 & 0.45 & 0.60  \\
    0.5 & 51.57 & 9.92  & 27.14 & 0.44 & 0.60  \\
    0.6 & 53.20 & 10.36 & 26.55 & 0.44 & 0.61   \\
    0.7 & 52.90 & 10.03 & 26.51 & 0.44 & 0.61 \\
    0.8 & 53.73 & 10.15 & 25.55 & 0.43 & \textbf{0.61} \\
    \bottomrule
\end{tabular}
\end{table}

\begin{table}[t!]
  \centering
  \small
  \setlength{\tabcolsep}{.9mm} 
  \caption{Comparison between the improved MDTv2 and original MDT. 
    MDTv2-S/2 are trained with 600k iterations. 
  }
  \vspace{-8pt}
  \label{tab:v1v2}
  \begin{tabular}{lcccccc} \toprule
    & Adan & MinSNR & Wider & Macro- & Decoder & \multirow{2}{*}{FID$\downarrow$} \\	
    & Opt.~\cite{xie2022adan} & \cite{hang2023efficient} & mask ratio & structure  & position	  & \\	\midrule
    MDT &  &  &  &  &  &  50.26 \\ \midrule
    \multirow{4}{*}{Improv.}
    & \checkmark &  &  &  &  & 49.71\\
    & \checkmark & \checkmark &  &  &  & 45.06 \\
    & \checkmark & \checkmark & \checkmark  &  &  & 43.82 \\
    & \checkmark & \checkmark & \checkmark & \checkmark &  & 38.41 \\ \midrule
    MDTv2 & \checkmark & \checkmark & \checkmark & \checkmark & \checkmark & 35.67  \\\bottomrule
  \end{tabular}
\end{table}

\myPara{Block number in side-interpolater.}
We compare the performance of different numbers of blocks in the side-interpolater
in~\tabref{tab:num_block_si}.
The default setting of 1 block achieves the best performance,
and the FID worsens with an increase in block number.
This result is consistent with our motivation that side-interpolater
should not learn too much information other than interpolating the masked representations.

\myPara{Positional-aware enhancement.}
To further release the potential of mask latent modeling,
we enhance the DiT baseline with stronger positional awareness ability,
\ie learnable positional embeddings and the relative positional bias in basic blocks.
\tabref{tab:pos_embed_si} shows the positional embeddings in side-interpolater
improves the FID from 51.58 to 50.26, indicating the positional embedding
is vital for the side-interpolater.
Also, enables the training of positional embeddings brings
the gain in FID as revealed in~\tabref{tab:learn_pos}.
In~\tabref{tab:rel_bias},
the relative positional bias in the basic blocks significantly improves
the FID from 53.56 to 50.26, showing
the relative positional modeling ability is essential
for diffusion models to obtain the contextual representation ability
and generate high-quality images.
Therefore, the positional awareness ability in diffusion model structure
is required to accompany the masked latent modeling,
playing a key role in improving performance.

\myPara{Comparison of VAE decoders.}
To ensure fair comparisons with DiT~\cite{peebles2022scalable}, we use both the MSE and EMA versions
of pretrained VAE decoders\footnote{MSE and EMA versions of VAE models are downloaded in \url{https://huggingface.co/stabilityai/sd-vae-ft-mse}
and \url{https://huggingface.co/stabilityai/sd-vae-ft-ema}.} for image sampling.
\tabref{tab:emavsmse} shows that
the EMA version has slightly better performance than the MSE version.
Except for the results in Table 1 of the manuscript that uses the EMA VAE decoder,
we use the MSE VAE decoder by default.

\begin{table}[t!]
  \centering
  \setlength{\tabcolsep}{2mm} 
  \small
  \caption{Comparison between the EMA and MSE version of VAE decoders.
    -G denotes the results with classifier-free guidance.
  }\vspace{-8pt}
  \label{tab:emavsmse}
  \begin{tabular}{lcccccc}
      \toprule
      Method	& Decoder &  FID$\downarrow$ & sFID$\downarrow$ & IS$\uparrow$ & Prec.$\uparrow$ & Rec.$\uparrow$ \\	\midrule
      MDT & MSE  & 6.65  & 5.07 & 129.47 & 0.72 & 0.63 \\
      MDT & EMA  & 6.46  & 4.92 & 131.70 & 0.72 & 0.63 \\
      \midrule
      MDT-G & MSE  & 2.14  & 4.45 & 259.21 & 0.82 & 0.59 \\ 
      MDT-G & EMA  & 2.02  & 4.46 & 263.77 & 0.82 & 0.60 \\ 
      \bottomrule
  \end{tabular}
\end{table}

\begin{table}[t!]
  \centering
  \small
  \setlength{\tabcolsep}{3.7mm} 
  \caption{Effect of the macro-structure designs in MDTv2.}
  \vspace{-8pt}
  \label{tab:structure_improvement}
  \begin{tabular}{llccccc}
      \toprule
      &Settings   &  FID$\downarrow$  \\	\midrule
      MDT & &  43.82 \\ \midrule
      \multirow{3}{*}{MDTv2} &
      + Encoder with long-shortcuts. & 41.12 \\
       & + Decoder with dense input-shortcuts. & 38.41 \\  
       & + Change to 6 decoders. & 35.67 \\
       
      \bottomrule
  \end{tabular}
\end{table}

\subsection{Ablation on MDTv2}
This section presents an ablation study focusing on the enhancements achieved by MDTv2 compared to the original MDT. 
We also report the results of MDTv2-S/2 model and use FID-50k as the evaluation
metric unless otherwise stated.
The summary of improvements in MDTv2 is presented in~\tabref{tab:v1v2}, highlighting how its new structure and enhanced training schemes significantly enhance performance.

\begin{table}[!t]
  \centering
  \small
  \setlength{\tabcolsep}{1.7mm} 
  \caption{Effect of position of side-interpolater
  in MDT v1/v2. 
    MDT-S/2 models contain 12 blocks.
    600k and 1500k denote the training iterations.
  }\label{tab:decoder_pos}
  \vspace{-8pt}
  \begin{tabular}{ccccccc}\toprule
    \multirow{2}{*}{Decoder pos.} & \multicolumn{3}{c}{FID$\downarrow$} \\ \cline{2-4}
    &  MDT (600k) & MDTv2 (600k) & MDTv2 (1500k) \\	\midrule
    Last0  & 51.05 &  -  & - \\
    Last2 & \textbf{50.26} & 38.41 & 33.20    \\
    Last4  & 51.67 & 36.53  & 31.05 \\
    Last6  & 52.64 &  \textbf{35.67} & \textbf{30.81} \\
    Last8  & -    &   37.93 & 33.42\\
    \bottomrule
  \end{tabular}
\end{table}

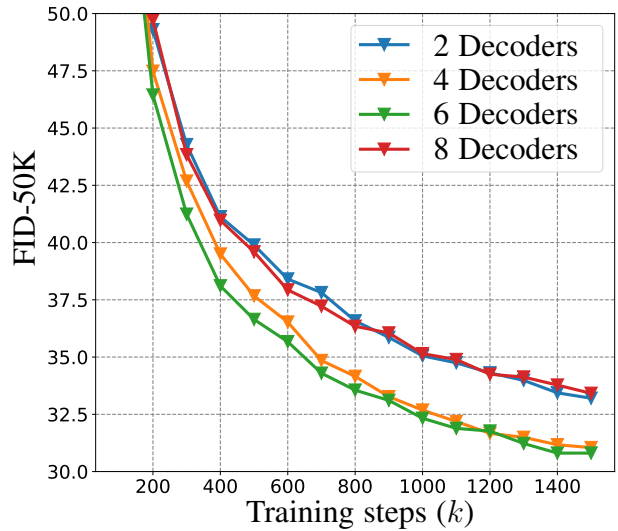
\begin{figure}[t]
  \footnotesize
  \centering
  \begin{overpic}[width=0.85\linewidth]{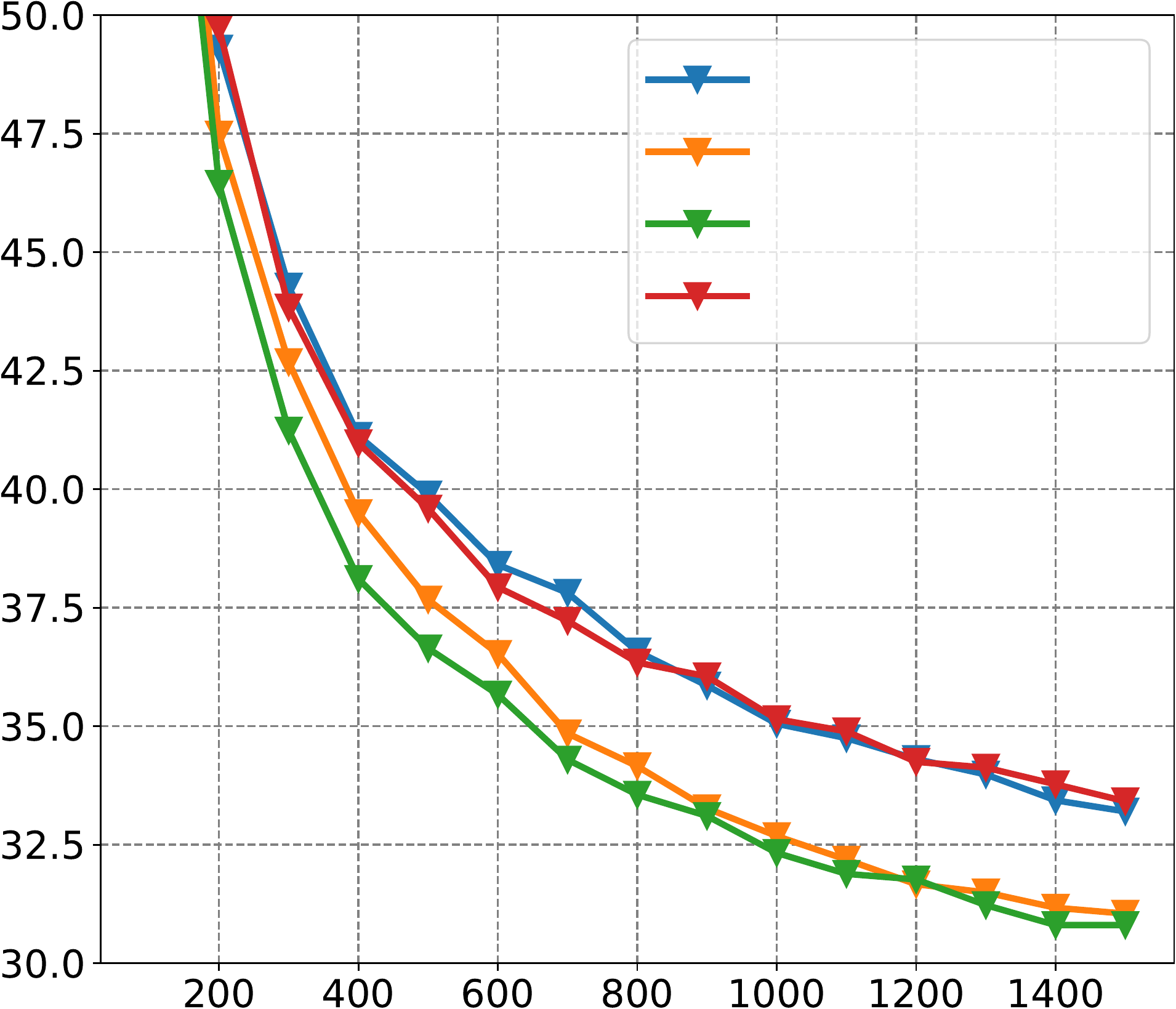}
    \put(-6,40){\large \rotatebox{90}{FID-50K}}
    \put(35,-5){\large Training steps ($k$)}
    \put(68,77){\large2 Decoders}
    \put(68,71){\large4 Decoders}
    \put(68,65){\large6 Decoders}
    \put(68,59){\large8 Decoders}
  \end{overpic}\vspace{10pt}
  \caption{Training convergence comparison of MDTv2
  under different numbers of decoder blocks.
  FID-50K is evaluated.
  }\label{fig:decoder_num}
\end{figure}

\myPara{Macro-structure in MDTv2.}
MDTv2 is designed with a macro-structure that includes an encoder featuring long-shortcuts and a decoder equipped with dense input-shortcuts.
As demonstrated in~\tabref{tab:structure_improvement},
enhancing the encoder with long-shortcuts improves the FID score from 43.82 to 41.12. 
The introduction of dense input-shortcuts in the decoder further reduces the FID score significantly by 2.71.
Unlike the original MDT, where noised patches are removed, the dense input-shortcuts enable MDTv2 to more accurately predict the noise values in masked patches, as the decoder now has access to the noised patches. 
Additionally, expanding the number of decoder blocks from 2 to 6 results in a stable performance improvement.

\myPara{Wider Masking ratio.} 
The masking ratio determines the number of input patches that
can be processed during training.
We give the comparison of using different masking ratios
in~\tabref{tab:maskratio}.
The best masking ratio for MDT-S/2 is 30\%,
which is quite different from the masking ratio used
for recognition models, e.g.   75\%
masking ratio in MAE~\cite{he2022masked}.
We assume that 
the image generation requires learning
more details from more patches for high-quality synthesis,
while recognition models only need
the most essential patches to infer semantics.
Compared to MDT, we demonstrate in \tabref{tab:v1v2} that MDTv2 gains advantages from employing a broad masking ratio between 0.3 and 0.5. This suggests that using variable masking ratios enables the model to learn diverse contextual representations.

\myPara{Side-interpolater position.}
To meet the high-quality image generation requirements of the diffusion model,
the side-interpolater is placed
in the middle of the network instead
of the end of the network in recognition models~\cite{he2022masked, bao2021beit}.
\tabref{tab:decoder_pos}
presents the comparison of placing 
the side-interpolater
at different positions of the MDTv2-S/2 model
with 12 blocks.
The results show that placing the side-interpolater 
in the middle of the model
achieves the best FID score,
whereas placing it at the end of the network like recognition models 
impairs the performance.
Placing the side-interpolater at
the early stages of the network also harm the performance,
indicating the mask latent modeling 
is beneficial to most stages in the diffusion models.
For MDT, placing
it before the last two blocks performs best,
while MDTv2 requires deeper decoders for better performance.
For MDTv2, with 600k training iterations, 
6 decoder blocks outperform those with 4 decoder blocks. 
However, when trained for an extended duration, 
e.g. 1500k iterations, 
both 4 and 6 decoder configurations achieve 
comparable performance.
We present the convergence curve for varying numbers of decoder blocks in \figref{fig:decoder_num}. The model configured with 6 decoders converges more rapidly in the early stages of training. However, this performance advantage diminishes over extended training periods.
While
both shallow decoder (2 blocks) and exceedingly deep decoder (8 blocks)
result in suboptimal convergence speed.

\section{Conclusion}
This work proposes a masked diffusion transformer
to enhance the contextual representation
and improve the relation learning among image semantics for DPMs.
We introduce an effective mask latent
modeling scheme into DPMs and also accordingly designs an asymmetric
masking diffusion transformer structure. 
Experiments show that our masked diffusion transformer enjoys higher performance on image synthesis and largely improves the learning progress during training,
achieving the new SOTA for image synthesis on the ImageNet dataset.

\section*{Acknowledgement.}
This research was supported by the NSFC (NO. 62225604) and
the Fundamental Research Funds for the Central Universities 
(Nankai University, 070-63233089).
The Supercomputing Center of Nankai University supports computation.

{\small

\bibliographystyle{ieee_fullname}
\bibliography{egbib}
}
  
\end{document}